\documentclass[runningheads]{llncs}

\usepackage{graphicx}
\usepackage{epsfig}
\usepackage{graphicx}
\usepackage{amsmath}
\usepackage{amssymb}
\usepackage{booktabs}
\usepackage{multirow}
\graphicspath{ {./images} }
\usepackage{epstopdf}

\usepackage{bm}

\usepackage{subcaption}

\usepackage[dvipsnames,svgnames,table]{xcolor}
\usepackage{longtable, makecell, multirow}
\usepackage{multirow}
\usepackage{sidecap} % side captions
\usepackage{tabularx}
\usepackage{floatrow}
\newfloatcommand{capbtabbox}{table}[][\FBwidth]
\usepackage{amsfonts}
\usepackage{mathrsfs}
\usepackage{gensymb}
\usepackage{mathtools}
\usepackage{upgreek}

\usepackage{leftindex}%mathtools}%amsmath}
\newcolumntype{?}{!{\vrule width 1pt}}
\usepackage[labelfont=bf]{caption}
\usepackage{subcaption}
\newdimen\imageheight % goes into the preamble

\usepackage{epstopdf}

\usepackage{tabularx}

\newcommand{\etal}{\textit{et al}.}

\begin{document}

% \title{Improving the Reliability of CNNs for Magnetic Resonance Spectral Modeling}%\thanks{Supported by organization x.}}
\title{Improving the Precision of CNNs for Magnetic Resonance Spectral Modeling}%\thanks{Supported by organization x.}}
\titlerunning{Improving Precision of CNNs}

\author{John LaMaster\inst{1}\orcidID{0000-0002-2149-771X} \and
Dhritiman Das\inst{2}\orcidID{0000-0001-6627-0618} \and
Florian Kofler\inst{1}\orcidID{2222--3333-4444-5555} \and
Jason Crane\inst{3}\orcidID{2222--3333-4444-5555} \and
Yan Li\inst{3}\orcidID{2222--3333-4444-5555} \and
Tobias Lasser\inst{1}\orcidID{0000-0001-5669-920X} \and
Bjoern H Menze\inst{4}\orcidID{0000-0003-4136-5690}}

\authorrunning{J. LaMaster et al.}
\institute{Munich Institute of Biomedical Engineering, Technical University of Munich, Munich, Germany \email{john.t.lamaster@gmail.com}\and
Massachusetts Institute of Technology, Cambridge, MA, USA \and
University of California, San Francisco, San Francisco, CA, USA \and
University of Zurich, Zurich, Switzerland}

\maketitle              % typeset the header of the contribution

\begin{abstract}
    Magnetic resonance spectroscopic imaging is a widely available imaging modality that can non-invasively provide a metabolic profile of the tissue of interest, yet is challenging to integrate clinically. One major reason is the expensive, expert data processing and analysis that is required. Using machine learning to predict MRS-related quantities offers avenues around this problem, but deep learning models bring their own challenges, especially model trust. Current research trends focus primarily on mean error metrics, but comprehensive precision metrics are also needed, e.g. standard deviations, confidence intervals, etc.. This work highlights why more comprehensive error characterization is important and how to improve the precision of CNNs for spectral modeling, a quantitative task. The results highlight advantages and trade-offs of these techniques that should be considered when addressing such regression tasks with CNNs. Detailed insights into the underlying mechanisms of each technique, and how they interact with other techniques, are discussed in depth.

% The abstract should briefly summarize the contents of the paper in
% 15--250 words.

\keywords{MRSI \and Spectral Fitting \and Quantification \and Deep Learning \and Reliability.}
\end{abstract}
\section{Introduction} \label{sec:Intro} 
Magnetic resonance spectroscopic imaging (MRSI) is a non-invasive, in-vivo clinical imaging modality commonly used to investigate the metabolic profile of tissue in order to evaluate various diseases. Metabolite nuclei interact with the radiofrequency pulses to generate a signal that can be quantitatively analyzed to characterize the metabolic profile of the tissue of interest. In order to generate reliable concentration maps, accurate metabolite quantification methods are required. This is achieved through various model-fitting tools such as QUEST \cite{Ratiney2005}, jMRUI \cite{Stefan2009}, TARQUIN \cite{Wilson2011}, and Osprey \cite{Oeltzschner2020}. Of these, the gold standard is a non-linear optimization method called the LCModel \cite{Provencher1993}. Unfortunately, this suffers from a number of bottlenecks including: 1) long fitting times; 2) manual parameter tuning requiring expert input; and 3) high quantification errors for noisy data. Both the time and expertise for acquiring, processing, and analysis hinder clinical adoption of MRSI, which is often limited to qualitative analyses. 

Model-based algorithms \cite{Garcia2010,Henry1997,Ravanbakhsh2015} were initially the standard approach in MRSI for spectral fitting and metabolite quantification. Das et al., in \cite{Das2017}, first demonstrated the use of traditional machine-learning methods for quantification, followed by deep-learning techniques in \cite{Gurbani2019,Hatami,Lee2019}. All DL regression tasks require models to be both accurate and reliable. Most research reports the mean model accuracy, but relevant precision metrics are largely neglected even though they are simple to calculate.

There are two primary contributions of this work: 1. It highlights the importance of including precision metrics in error characterizations; and 2. It provides and characterizes the effects of various techniques for improving the precision of CNNs for metabolite quantification of MRS data. Sec.\ref{sec:Experiments} presents a compiled ablation study that evaluates the effects of selected techniques on model accuracy, precision, and stability. The reported techniques can be incorporated into existing architectures, or used when designing new models. They can be combined with other techniques, such as data augmentation strategies or task-specific optimization routines, to further improve model performance.

\section{Methodology}\label{sec:Methods}
\subsection{Architecture} \label{subsec:Architecture}
\begin{figure}
    \centering
    \begin{subfigure}[b]{0.49\textwidth}
        \centering
        \includegraphics[width=0.75\textwidth, keepaspectratio]{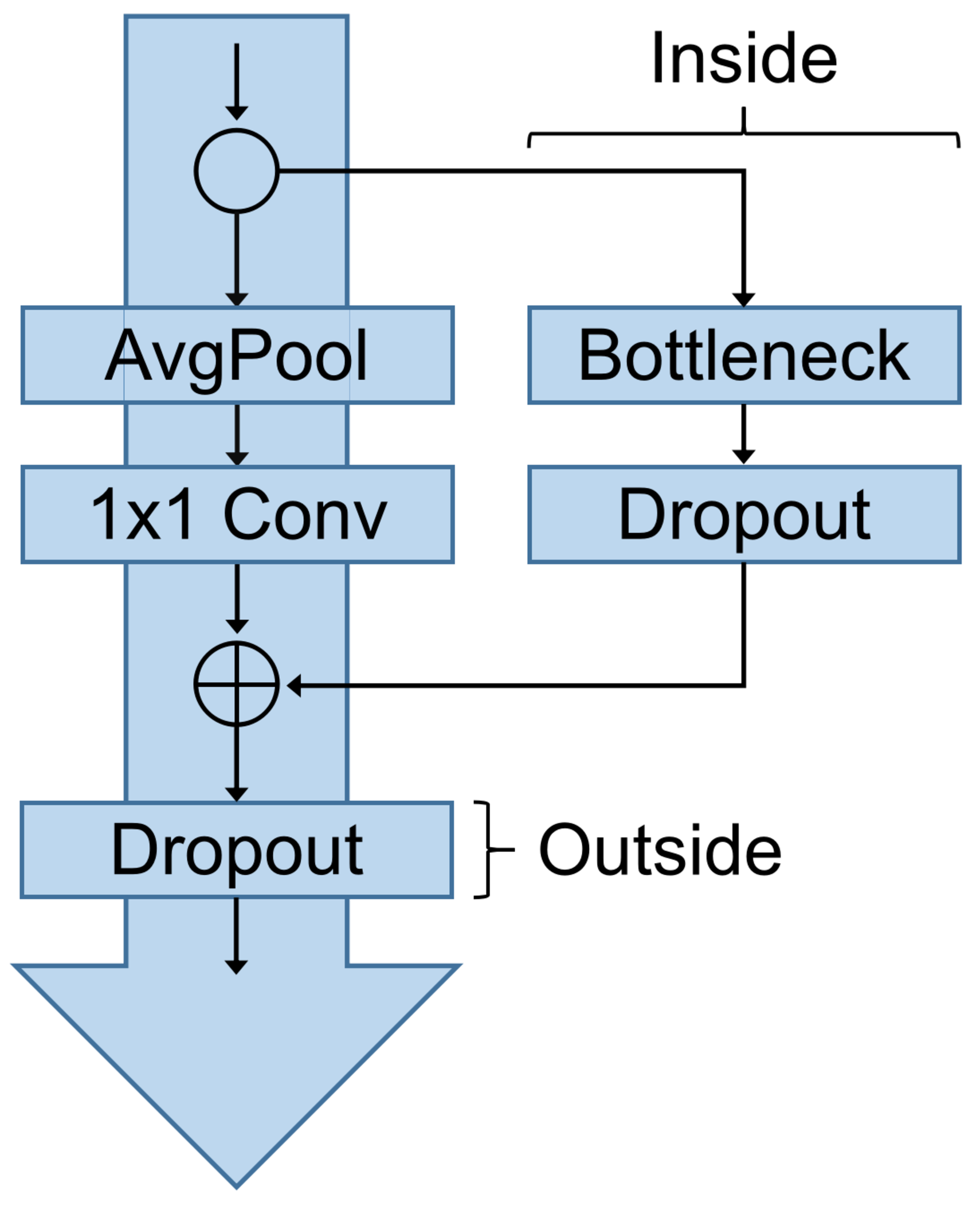}
        \caption{Residual block architecture.}%Template architecture of a convolutional layer.}
        \label{fig:ConvLayer}
    \end{subfigure}
    \begin{subfigure}[b]{0.49\textwidth}
        \centering
        \includegraphics[width=0.60\textwidth, keepaspectratio]{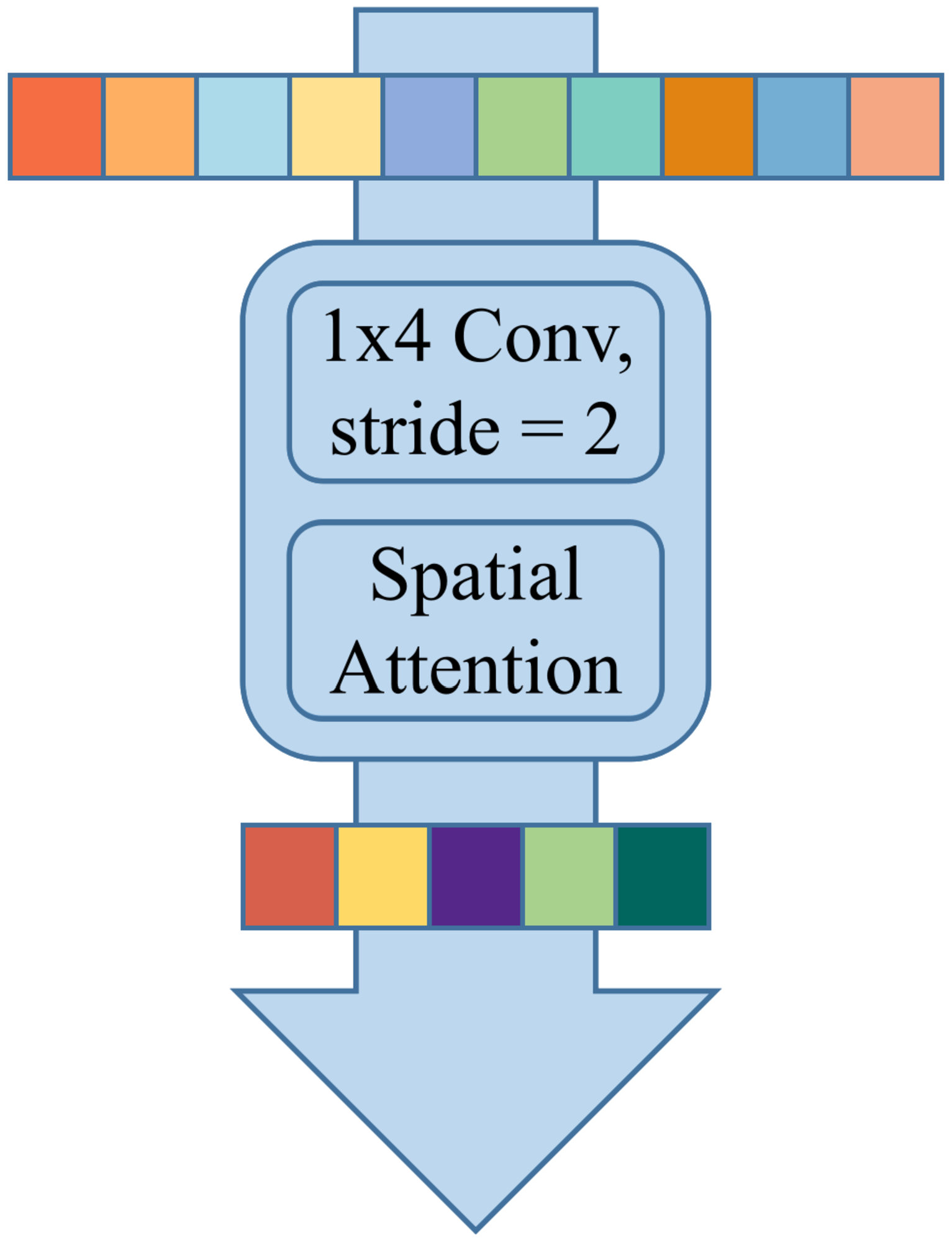}   
        \vspace{3.0mm}
        \caption{Spatial feature condenser module.}%to downsample excessively large feature maps.}
        \label{fig:Feature Condenser}
    \end{subfigure}
    \caption{Reference module architectures.}
    \label{fig:architectural blocks}
\end{figure}

A ResNet50 is used as the vanilla architecture with several state-of-the-art upgrades. The base is the \emph{PreAct-ResNet50\_v1.5}. As suggested by \cite{Hatami}, ReLU activation functions are replaced in the first half of the network with CReLU as described in \cite{Shang2016}. Next, the downsampling step and the skip connection were updated with ResNet-b and -d from \cite{He2019}. Fig.\ref{fig:ConvLayer} shows the template used for the convolutional layers. \textit{Inside} and \textit{Outside} refer to the modules' location with respect to the block's skip connection. Standard spectral lengths in MRS are 1024 points. This work crops and resamples the data to length 512, which is still larger than the standard implementation of 224. Further downsampling would risk losing features important for the quantitative nature of this work. Pooling larger features before the fully-connected (FC) layer destabilizes the networks. Therefore, a spatial feature condenser is needed before the FC layer to reduce the feature size before pooling. The condenser is a series of blocks, shown in Fig.\ref{fig:Feature Condenser}, containing a channel-wise, strided convolution, and an optional spatial attention gate that learn to downsample the features. 

\subsection{Dropout Techniques}\label{subsec:Dropout}
This work focuses on structured dropout techniques as a means to improving the informativeness and stability of computer vision CNNs. 

\subsubsection{Global dropout} The first technique is a data dropout technique called \textit{dropCluster} (\textit{dC}) \cite{Chen2020}. This has a global effect because it drops entire contiguous features at the beginning of the network right after the stem. It uses a feature agglomeration clustering technique to group intra-channel features. A stable feature representation must be learned before the clustering can work. The dropout rate for each cluster is then modulated based on its size and linearly increases from zero to \(p\) over a prolonged warm-up phase, typically the remainder of training. Following the original implementation, this is only activated once a steady feature representation is learned. An activation epoch of 10 was determined experimentally. 

\subsubsection{Local dropout} 
The remaining techniques are local and related. The first is \textit{Feature Alpha Dropout} (\textit{FAD}) which randomly drops entire channels. Instead of replacing the dropped channels with zero, it uses the negative saturation value from the SELU activation and then applies a transform to maintain the original mean and variance. The second method is \textit{weighted Feature Dropout} (\textit{wFD}) which scores the channels according to their level of activation and then drops those with the highest activations. It is based on Hou \etal's method in \cite{Hou2019}. The scoring system calculates the channel-wise means which are converted to the logarithmic scale and normalized to $[0,1]$. All values below the hyperparameter $q=0.90$ were set to zero. The resultant values were multiplied by the drop probability, $p$, to calculate the effective dropout rate for each channel. A Bernoulli random variable is used to select which channels to drop which prevents the network from always dropping the same channels and collapsing. The third method, \textit{weighted Feature Alpha Dropout} (\textit{wFAD}), is a novel implementation that combines the weighting scheme from the \textit{wFD} with the alpha implementation of \textit{FAD}. 

Preliminary experiments confirmed \cite{Zoph2018}'s findings that such strong regularizers require annealing strategies to prevent instability and model divergence. This work used a linear strategy scaling $p$ starting in epoch 10 from 0 to $p$ over the length of training. Gradually omitting features forces the network to identify both richer and more complementary representations. As \cite{Zoph2018} suggested, the ratio of dropout rates for ResNet layers $1-4$ were set to [1$p$, 2$p$, 3$p$, 4$p$] with $p$ being the reported dropout rate. Preliminary experiments found that values greater than 0.10 degraded performance for this task. The effective dropout rate for each layer is defined as follows:
\begin{equation}
    \label{eqn:effective dropout rate} 
    p_{eff} = p_{max}\textnormal{ }\lambda_{sched}\textnormal{ }\hat{s}, \textnormal{ where } s = \textnormal{log} \left( \frac{\bar{x}}{\sum_{i}\bar{x}_{i}}\right)
\end{equation}
where \(p_{max}\) is the maximum drop rate, \(\lambda_{sched}\) scales \(p_{max}\) according to the schedule, \(\bar{x}\) is the channel-wise mean of the data, \(\hat{s}\) is the unit normalization of the channel ratings. $p_{eff}$ was used to define the Bernoulli random variable for each channel which then selected the channels to drop.

\subsection{Task complexity}
The dataset, described in Sec. \ref{subsubsec:MRS Data Simulation}, is defined by 14 variables that are a mixture of dependent, independent, linear, and non-linear variables. A simplified dataset was defined using the same physics model but only 7 variables: PCh, Cre, NAA, MM, Lip, T2$^*$, and SNR. Then a more complex dataset was created using a zero-order phase offset and a Voigt lineshape, consisting of one Lorentzian value (D) per metabolite and one Gaussian value (G) per spectrum. This model used a total of 26 variables: PCh, Cre, NAA, Glx, Ins, GPC, Tau, MM, Lip, D, G, Phi0, SNR, and the 5 baselines from the original model. The latter model has more dependent and independent variables and will provide a more challenging learning task. A baseline ResNet50 will be compared against the best performing model identified in Table \ref{tab:Ablation Results - dropout}.

\subsection{Implementation} \label{subsec:Implementation}
\subsubsection{MRS data simulation}\label{subsubsec:MRS Data Simulation}
The learning task in this work is a supervised, multivariate regression that learns the parameters of the physics model used to simulate the training data. This physics model comes from Das \etal \cite{Das2017} and is shown in Eq. \ref{eqn:Das model} below: 
\begin{equation}\label{eqn:Das model}
    S(\omega) = \mathcal{F} \left \{\int p(\omega)e^{-i\Phi}e^{-t/T_2^*}d\omega \right \}
\end{equation}

where $p(\omega)$ is the modulated basis function, $\Phi$ includes the zero- and first-order phase offsets, and $t/T_2^*$ applies a Lorentzian lineshape. This is a linear combination model that modulates metabolite basis functions, \(\theta_M\), and then adds a lineshape profile, noise, and a spectral baseline offset. As in \cite{Das2017}, there are 5 metabolite basis functions- PCh, Cre, NAA, Glx (Gly+Gln), and Ins- plus 5 baseline basis functions from \cite{Das2018a} that result in 14 total parameters. The 5 baseline basis functions are randomly selected. The simulated Gaussian noise is constrained to a SNR range of [5,30]. All variables are sampled from uniform distributions. After simulation, the data is cropped to the ppm range \([0.2, 4.2]\) and resampled to length 512. The training dataset consists of 125,000 spectra split $80/20$ for training and validation.

\paragraph{Quantification} In MR spectroscopy, metabolites are quantified through a process known as quantification. This involves modeling various spectral components using an iterative, non-linear solver. Each metabolite in the data is visible through one or more spectral peaks in the frequency domain. These peaks exist at known frequencies and their peak height, or area, represents their respective concentration. To quantify this, a pre-simulated basis function for each metabolite is iteratively scaled and broadened until it matches the observed data. DL approaches to this task aim to replace the iterative, non-linear solver with a single-shot neural network.

\subsubsection{Metrics}\label{subsubsec:Metrics}% and Loss Function} 
The primary error metric is the mean absolute percent error (MAPE) of the metabolite quantities and is reported with its standard deviation (STD), which is important to evaluate the reliability of the model. To assess the regression, the coefficient of determination, \(r^2\) is reported for the cumulative set of metabolites. The p-value is omitted here because it is sample size-dependent and this work uses sample sizes large enough that the p-values converge to zero. However, reporting it would be important when evaluating data sets with small sample sizes. A new generalized consistency metric, \(\bar{S}\), is also reported. \(\bar{S}\) is defined in Eqn. \ref{eqn:smoothness metric} as the variance of the second derivative of the metric's temporal curve. This is a way to measure the smoothness of a metric's landscape which is reflective of the stability of the model's features. 
\begin{equation}\label{eqn:smoothness metric}
    \bar{S}_{i} = \textbf{var} \left( \int_0^l \frac {d^2m}{dx^2} dx \right)
\end{equation}

where \textit{m} is the training or validation curve of the metric being monitored, \textit{x} is the x-axis of the curve in units of epochs, the integral is approximated by a Riemann sum, \textit{l} is the number of epochs, and \textit{i} is the metric being evaluated. \(\bar{S}\) can be calculated for any temporally monitored metric for any given parameter because it is agnostic to the metric's equation. When a model performs very consistently over the training or validation period, the metric's curve will be very smooth and the variance of the curve's local second derivative will be lower, indicating that the magnitude of the variance is inversely proportional to the temporal consistency of the model's performance for the given metric. A large \(\bar{S}\) indicates erratic performance and unstable features. Low \(\bar{S}\) and a small STD indicate the model has learned more stable and richer feature representations. High \(r^2\) and low MAPE scores also indicate more informative features were learned.

\subsubsection{Dropout rates}\label{subsubsec:dropout rates}
The dropout rate is an important hyperparameter in deep learning. In standard, qualitative tasks with the original dropout implementation, $p=0.1$ is the standard rate. As mentioned above, structured dropout techniques are very strong regularizers and require greater consideration than standard approaches. Therefore, each technique was tested using three separate dropout rate- 0.10, 0.05, and 0.025- for a total of 30 experiments for the original ablation study. The complete table will be published with the repository. The best performers for each technique were compiled into Table \ref{tab:Ablation Results - dropout}.

\subsubsection{Loss function}\label{subsubsec:Loss}
All experiments in this work are trained using a standard mean squared error (MSE) loss calculated on the entire output and the parameters of the spectral components both individually and grouped as follows: metabolites, line broadening factor, noise, and baseline coefficients. Each of those losses is accompanied by a corresponding weight, \(\lambda_{\theta}\), described in Eqn. \ref{eqn:realtive lambdas}. The parameter-specific lambdas are calculated relative to their validation regression metrics \(r_{\theta}\) and \(r^2_{\theta}\), the consistency metric \(\bar{s}_{\theta}\), and a simple epoch penalty to discourage sub-optimal performance plateaus. This allows for autonomous optimization and encourages disentanglement of the spectral components.

\begin{equation}
    \begin{gathered}
    \label{eqn:realtive lambdas}
        \textnormal{pen}_{epoch} =  epoch / 100\\
        \textnormal{value} = \left((1 - r) + (1 - r^2) + \bar{S}_i \right)  \left(10 + \textnormal{pen}_{epoch}\right)\\
        \lambda_{\theta} = max(\textnormal{value}; \textnormal{pen}_{min} + \textnormal{pen}_{epoch} + \bar{S}_i)
    \end{gathered}
\end{equation}

\subsubsection{Training}
\label{subsubsec:Training}
All trials are trained with a standard Adam optimizer using a learning rate of $lr=0.001$ and a batch size of 250. The structured dropout techniques are activated after 10 epochs and then warmed-up for the remainder of training. All models in the ablation studies are trained for 100 epochs. This was selected because the validation curves show that the largest amount of learning generally occurs in the first 20 epochs, then performance begins to plateau. The code is written in PyTorch 1.4.0 and all models were trained on a single 12GB Nvidia TITAN Xp GPU.

\section{Experiments}\label{sec:Experiments}
The effects of the four structured dropout techniques described above are detailed in Table \ref{tab:Ablation Results - dropout}. The location, indicated by a subscripted \textit{I} or \textit{O}, indicates placement inside or outside of the residual blocks, i.e. before or after the skip connection. 

\subsection{Cluster dropout} \textit{dC} did not show much effect on its own and results were the same for all dropout rates. In theory, \textit{dC} should force the network to pay attention to more features of the input and make better use of the collection of features provided by the network stem. In practice, this seems to fail likely because it's dropout rate is scaled down proportional to the size of the given feature, meaning that it preferentially drops very small features regardless of their relevance. This is supported by the poorer MAPE and STDs and the improvement in temporal consistency. Remember that the weighting scheme does not use a Bernoulli distribution meaning that it consistently drops the same features, preventing the network from learning to use them. This combined with the worse MAPE values indicate that a non-trivial amount of the smaller features are in fact informative. The large improvement in \(\bar{s}\), however, implies that other small features are analogous to noise in the training distribution and eliminating them stabilizes performance.

\begin{table*}[t]
    \centering
    \begin{tabular}{|cl|*{6}{c|}} \cline{3-8}
        \multicolumn{2}{c|}{} & \textbf{drop prob} & \textbf{Epoch} & \textbf{MAPE} & \textbf{STD} & $\boldsymbol{r^2} \thead{\uparrow}$ & $\boldsymbol{\bar{s}} \downarrow$ \\\hline\hline
        \multirow{1}{*}[0em]{I} & ResNet50 & - - & 97 & 22.51 & 13.10 & 0.96 & 71.27 \\\hline\hline
        \multirow{6}{*}[-.55em]{II} & dropCluster & 0.10 & 88 & 23.11 & 14.46 & 0.96 & 37.23 \\\cline{2-8}
        & FAD$_O$ & 0.025 & 29 & 17.78 & 4.04 &  0.94 & 3.75 \\
        & wFD$_O$  & all & na & na & na & na & na \\
        & wFAD$_O$  & 0.05 & 97 & \textbf{16.30} & \textbf{3.14} & 0.95 & \textbf{1.86} \\\cline{2-8}
        & FAD$_I$ & 0.05 & 71 & \textbf{19.71} & \textbf{8.70} & 0.95 & \textbf{16.34} \\
        & wFD$_I$ & 0.05 & 6 & 43.13 & 33.19 & 0.96 & 252.59 \\
        & wFAD$_I$ & 0.025 & 89 & 21.61 & 14.60 & 0.95 & 83.45 \\\cline{2-8}
        % & Best: in + out & & 92 & 24.71 & 22.13 & 0.96 & 0.0 & 45.35 \\
        \hline\hline
        \multirow{4}{*}[-.2em]{III} & dC, wFAD$_O$ & 0.10/0.05 & 89 & \textbf{17.64} & \textbf{5.01} & 0.95 & \textbf{9.24} \\
        & dC, FAD$_I$ & 0.10/0.05 & 81 & 18.89 & 7.41 & 0.94 & 25.94 \\\cline{2-8}
        & FAD$_I$, wFAD$_O$ & 0.025 & 90 & 17.71 & 2.32 & 0.94 & \textbf{1.12} \\%\cline{2-9}
        & dC, FAD$_I$, wFAD$_O$ & 0.10/0.025 & 88 & \textbf{16.24} & \textbf{2.13} & 0.94 & 5.14 \\
        \hline
    \end{tabular}
    \caption{Ablation Study of the dropout techniques grouped into three categories: baseline, individual, and combinations.}
    \label{tab:Ablation Results - dropout}
\end{table*}

\subsection{Feature alpha dropout} \textit{FAD} shows good and consistent improvements, especially with lower dropout rates both inside and outside the resblock. The $r^2$ values showed either a minimal decrease or no change. The MAPE values all improved. $p=0.025$ produced lower error values, but $p=0.05$ produced smaller standard deviations. These narrow STDs indicate that the learned features are more insensitive to variations between samples. This suggests that the models have learned more comprehensive feature representations. All dropout rates showed large improvements regarding \(\bar{s}\), especially after the skip connection. In both locations, \textit{FAD} greatly reduced the necessary number of training epochs.

\subsection{Weighted feature dropout} \textit{wFD} failed both inside and outside of the resblock. It quickly caused model collapse when applied after the skip connection. Inside the resblock, the skip connection was able to prevent model collapse but suffered severe performance degradation.

\subsection{Weighted feature alpha dropout} On the other hand, \textit{wFAD} showed great improvements. This module performs best in between convolutional layers rather than inside the resblock. It yielded a $5.21\%$ MAPE improvement, lowered the STD from by $13.10$ down to $3.14$, and improved the consistency metric $\bar{s}$ from $71.27$ to $1.86$. These results confirm the previous inference that narrower STDs indicate more comprehensive feature representations. This technique specifically drops the most activated channels, i.e. features. Therefore, it is the learning of additional complimentary features that reduces the STD. As Table 1 in the supplement shows, STDs of \textit{FAD} are consistently lower than \textit{wFAD}. This suggests that features with mid-level, and possibly low-level, activations are also important in building more comprehensive representations. As with \textit{FAD}, the dropout rate must be chosen carefully. Again in Supplement Table 1, the STDs and \(\bar{s}\) both improve with higher dropout rates but at the expense of the accuracy and $r^2$. \(p_{\textnormal{wFAD}}\) is therefore a very important hyperparameter.

\subsection{Combinations} 
Certain combinations of dropout inside and outside the resblocks produced STDs smaller than either one individually. This combination reduced the error by $4.80\%$, the STD by $10.78$, and \(\bar{s}\) from 71.27 to 1.12. Such strong regularization did reduce $r^2$ by 0.02. When combining the best performing inside and outside techniques with \textit{dC}, the results were mixed. When paired with \textit{wFAD$_O$}, the performance worsened while it improved when paired with \textit{FAD$_I$}. MAPE and STD both improved but $r^2$ and \(\bar{s}\) decreased. \textit{dC,FAD$_I$,wFAD$_O$} combined the benefits of all 3 methods. MAPE improved by $6.27\%$, STD decreased by $10.97$, and \(\bar{s}\) improved from 71.27 to 5.14. The only drawback is that $r^2$ decreased to 0.94. The results of these combinations of structured dropout techniques encourage more accurate (MAPE), more comprehensive (STD), and stable (\(\bar{s}\) feature representations.

\begin{table*}[t]
    \centering
    \begin{tabular}{|cl|*{6}{c|}} \cline{3-8}
        \multicolumn{2}{c|}{} & \textbf{drop prob} & \textbf{Epoch} & \textbf{MAPE} & \textbf{STD} & $\boldsymbol{r^2} \thead{\uparrow}$ & $\boldsymbol{\bar{s}} \downarrow$ \\\hline\hline
        \multirow{2}{*}[0em]{7-var} & Baseline & - - & 94 & 10.74 & 5.31 & 0.99 & 7.21 \\
        & $^*$Proposed & 0.10/0.025 & 83 & 15.89 & 6.02 & 0.98 & 27.68 \\\hline\hline

        \multirow{2}{*}[0em]{14-var} & Baseline & - - & 97 & 22.51 & 13.10 & 0.96 & 71.27 \\
        & $^*$Proposed & 0.10/0.025 & 88 & 16.24 & 2.13 & 0.94 & 5.14 \\\hline\hline

        \multirow{2}{*}[0em]{26-var} & Baseline & - - & 97 & 50.39 & 6.42 & 0.84 & 319.83 \\
        & $^*$Proposed & 0.10/0.025 & 23 & 47.44 & 2.00 & 0.83 & 36.54 \\\hline
    \end{tabular}
    \caption{Ablation study comparing the effect of the proposed dropout combination on regression tasks with varying levels of complexity.} \label{tab:Complexity}
\end{table*}

\subsection{Task Complexity}
In Table \ref{tab:Complexity}, one can see that the benefits of these techniques improves as the complexity of the regression task increases. The 14- and 26-variable datasets showed improvements in MAPE, STD, and $\bar{s}$. The 7-variable dataset showed similar scores for the STD and $r^2$, but showed degraded MAPE and $\bar{s}$. This lackluster performance is likely due, in part, to the ResNet50 being too large for the task. With each dataset, the $r^2$ value decreases slightly, but is still within 0.02 points of the baseline values.

\subsection{Limitations}
This work explored three different dropout probabilities, but did not perform a comprehensive optimization. The main objective was to observe the effects of the included techniques and that was achieved with a small group of dropout rates. Next, the nature of ablation studies requires the models to be deterministic and initialized with a fixed seed. While not necessary to prove their effect, a study testing various seeds would provide a more complete overview of the evaluated techniques for a given task. It should also be noted that the architecture was not tailored to the task. This work wanted to show that even unoptimized architectures can be improved with these techniques. Finally, a single annealing strategy was used for this entire work. While attention modules are capable of improving the MAPE scores, more complex annealing strategies should be able to overcome some of the current trade-offs between the various metrics.

\section{Conclusion}\label{sec:Conclusion}
In this work, four techniques falling into two categories were assessed for their impact on the error, standard deviation, coefficient of determination, and performance consistency. Many of these techniques have trade-offs in the quantitative setting. \textit{FAD} and the new \textit{wFAD} are simple and highly effective methods of narrowing these error ranges and stabilizing model performance. Table \ref{tab:Complexity} shows that these techniques are especially effective with more complicated learning tasks. 

Most importantly, this work shows the necessity of reporting precision metrics, such as standard deviations or confidence intervals, for regression tasks. The ablation study presented shows that when unaddressed, the error ranges can be so large that they render the models useless. Demonstrating and improving model reliability is key for translating research into into clinical and scientific practice.

\bibliographystyle{splncs04}
\bibliography{mybibliography}

\begin{thebibliography}{10}
\providecommand{\url}[1]{\texttt{#1}}
\providecommand{\urlprefix}{URL }
\providecommand{\doi}[1]{https://doi.org/#1}

\bibitem{Chen2020}
Chen, L., Gautier, P., Aydore, S.: {DropCluster: A structured dropout for convolutional networks}. arXiv preprint  (2020), \url{http://arxiv.org/abs/2002.02997}

\bibitem{Das2017}
Das, D., Coello, E., Schulte, R.F., Menze, B.H.: {Quantification of metabolites in magnetic resonance spectroscopic imaging using machine learning}. Lecture Notes in Computer Science (including subseries Lecture Notes in Artificial Intelligence and Lecture Notes in Bioinformatics)  \textbf{10435 LNCS},  462--470 (2017). \doi{10.1007/978-3-319-66179-7\_53}

\bibitem{Das2018a}
Das, D., Davies, M.E., Chataway, J., Chandran, S., Menze, B.H., Marchall, I.: {Metabolite Quantification of 1H-MRSI spectra in Multiple Sclerosis: A Machine Learning Approach}. ISMRM  (2018)

\bibitem{Garcia2010}
Garcia, M.I.O., Sima, D., Nielsen, F., Himmelreich, U., {Van Huffel}, S.: {Quantification of in vivo magnetic resonance spectroscopy signals with baseline and lineshape corrections}. 2010 IEEE International Conference on Imaging Systems and Techniques, IST 2010 - Proceedings pp. 349--352 (2010). \doi{10.1109/IST.2010.5548503}

\bibitem{Gurbani2019}
Gurbani, S.S., Sheriff, S., Maudsley, A.A., Shim, H., Cooper, L.A.: {Incorporation of a spectral model in a convolutional neural network for accelerated spectral fitting}. Magnetic Resonance in Medicine  \textbf{81}(5),  3346--3357 (2019). \doi{10.1002/mrm.27641}

\bibitem{Hatami}
Hatami, N., Sdika, M., Ratiney, H.: {Magnetic resonance spectroscopy quantification using deep learning}. Lecture Notes in Computer Science (including subseries Lecture Notes in Artificial Intelligence and Lecture Notes in Bioinformatics)  \textbf{11070 LNCS},  467--475 (2018). \doi{10.1007/978-3-030-00928-1\_53}

\bibitem{He2019}
He, T., Zhang, Z., Zhang, H., Zhang, Z., Xie, J., Li, M., Services, A.W.: {Bag of Tricks for Image Classification with Convolutional Neural Networks}. Proceedings of the IEEE/CVF Conference on Computer Vision and Pattern Recognition (CVPR) pp. 558--567 (2019)

\bibitem{Henry1997}
Henry, E.R.: {The use of matrix methods in the modeling of spectroscopic data sets}. Biophysical Journal  \textbf{72}(2 I),  652--673 (1997). \doi{10.1016/S0006-3495(97)78703-4}

\bibitem{Hou2019}
Hou, S., Wang, Z.: {Weighted channel dropout for regularization of deep convolutional neural network}. In: 33rd AAAI Conference on Artificial Intelligence, AAAI 2019, 31st Innovative Applications of Artificial Intelligence Conference, IAAI 2019 and the 9th AAAI Symposium on Educational Advances in Artificial Intelligence, EAAI 2019. pp. 8425--8432 (2019). \doi{10.1609/aaai.v33i01.33018425}

\bibitem{Lee2019}
Lee, H.H., Kim, H.: {Intact metabolite spectrum mining by deep learning in proton magnetic resonance spectroscopy of the brain}. Magnetic Resonance in Medicine  \textbf{82}(1),  33--48 (2019). \doi{10.1002/mrm.27727}

\bibitem{Oeltzschner2020}
Oeltzschner, G., Z{\"{o}}llner, H.J., Hui, S.C., Mikkelsen, M., Saleh, M.G., Tapper, S., Edden, R.A.: {Osprey: Open-source processing, reconstruction \& estimation of magnetic resonance spectroscopy data}. Journal of Neuroscience Methods  \textbf{343},  1--27 (2020). \doi{10.1016/j.jneumeth.2020.108827}

\bibitem{Provencher1993}
Provencher, S.W.: {Estimation of metabolite concentrations from localized in vivo proton NMR spectra}. Magnetic Resonance in Medicine  \textbf{30}(6),  672--679 (1993). \doi{10.1002/mrm.1910300604}

\bibitem{Ratiney2005}
Ratiney, H., Sdika, M., Coenradie, Y., Cavassila, S., van Ormondt, D., Graveron-Demilly, D.: {Time-domain semi-parametric estimation based on a metabolite basis set}. NMR in Biomedicine  \textbf{18}(1),  1--13 (2005). \doi{10.1002/nbm.895}

\bibitem{Ravanbakhsh2015}
Ravanbakhsh, S., Liu, P., Bjordahl, T.C., Mandal, R., Grant, J.R., Wilson, M., Eisner, R., Sinelnikov, I., Hu, X., Luchinat, C., Greiner, R., Wishart, D.S.: {Accurate, fully-automated NMR spectral profiling for metabolomics}. PLoS ONE  \textbf{10}(5),  1--15 (2015). \doi{10.1371/journal.pone.0124219}

\bibitem{Shang2016}
Shang, W., Sohn, K., Almeida, D., Lee, H.: {Understanding and improving convolutional neural networks via concatenated rectified linear units}. 33rd International Conference on Machine Learning, ICML 2016  \textbf{5},  3276--3284 (2016)

\bibitem{Stefan2009}
Stefan, D., Cesare, F.D., Andrasescu, A., Popa, E., Lazariev, A., Vescovo, E., Strbak, O., Williams, S., Starcuk, Z., Cabanas, M., {Van Ormondt}, D., Graveron-Demilly, D.: {Quantitation of magnetic resonance spectroscopy signals: The jMRUI software package}. Measurement Science and Technology  \textbf{20}(10) (2009). \doi{10.1088/0957-0233/20/10/104035}

\bibitem{Wilson2011}
Wilson, M., Reynolds, G., Kauppinen, R.A., Arvanitis, T.N., Peet, A.C.: {A constrained least-squares approach to the automated quantitation of in vivo 1H magnetic resonance spectroscopy data.} Magnetic Resonance in Medicine  \textbf{65}(1),  1--12 (2011). \doi{10.1002/mrm.22579}

\bibitem{Zoph2018}
Zoph, B., Vasudevan, V., Shlens, J., Le, Q.V.: {Learning Transferable Architectures for Scalable Image Recognition}. Proceedings of the IEEE Computer Society Conference on Computer Vision and Pattern Recognition pp. 8697--8710 (2018). \doi{10.1109/CVPR.2018.00907}

\end{thebibliography}
\end{document}